\documentclass[runningheads]{llncs}
\usepackage{graphicx}

\usepackage[english]{babel}
\usepackage{babel,blindtext}
\usepackage[utf8]{inputenc}
\usepackage{amsmath}
\usepackage{amssymb}
\usepackage{mathtools}
\usepackage{amsfonts}
\usepackage{algorithm}
\usepackage{algpseudocode}
\usepackage{caption}
\usepackage{subcaption}

\usepackage{pgfplots}
\pgfplotsset{compat=1.18, width=10cm}
\newcommand{\quotes}[1]{``#1''}
\usepackage[margin=1.05in]{geometry}

\begin{document}
\title{Online Linear Regression Based on Weighted Average}
%
%
\author{Mohammad Abu-Shaira\inst{1}\and
Greg Speegle\inst{2}}
%
%
\institute{Baylor University, Waco TX 76798, USA 
\email{mohammad\_abu-shaira1@baylor.edu}\and
Baylor University, Waco TX 76798, USA
\email{greg\_speegle@baylor.edu}}
\maketitle              
\begin{abstract}
Machine Learning requires a large amount of training data in order to build 
accurate models. Sometimes the data arrives over time, requiring significant 
storage space and recalculating the model to account 
for the new data. On-line learning addresses these issues by incrementally modifying 
the model as data is encountered, and then discarding the data. In this study
we introduce a new online linear regression approach. Our approach combines newly arriving
data with a previously existing model to create a new model.
The introduced model, named \textit{OLR-WA} (OnLine Regression with Weighted Average) uses user-defined weights to provide flexibility in the face of changing data to bias the results in favor of old or new data. 
We have conducted 2-D and 3-D experiments comparing OLR-WA to a static model using the entire data set.
The results show that for consistent data, OLR-WA and the static batch model perform similarly and for varying data,
the user can set the OLR-WA to adapt more quickly or to resist change. 

\end{abstract}

\section{Introduction}
In Machine Learning, the conventional batch approach operates under the assumption that all data is accessible for every computation. This allows 
many benefits such as repeatedly accessing the data and cross-validation by leaving portions of the data out.
Furthermore, the batch learning approach assumes \cite{fontenla2013online}: 
\begin{enumerate}
    \item the whole training set can be accessed to adjust the model;
    \item there are no time restrictions, meaning we have enough time to wait until the model is completely trained; 
    \item the data distribution does not change; it is typically assumed to be independently and identically distributed (iid). After the model is calibrated, it can produce accurate results without the need for further adjustments. 
\end{enumerate}
\noindent

However, these assumptions limit the applicability of the batch approach \cite{fontenla2013online}. For example
consider the scenario of a machine learning model that is trained to predict stock prices. The model is initially trained on historical stock market data and is used to make predictions about future stock prices. However, as time passes, the stock market changes. For example, the economy can go through a recession or a period of high inflation, new companies can go public, and old companies can go bankrupt. These changes in the stock market mean that the initial training set used to train the model is no longer valid. In order to continue making accurate predictions, the model must be updated to adapt it to the new conditions. Another scenario is in streaming environments where predictions are required at any given moment during execution. In such cases, the batch model must be recreated each time a prediction is required \cite{fontenla2013online}. For example, consider the scenario of a traffic management system in a smart city. In this case, the system needs to provide real-time predictions for traffic flow and congestion levels at different locations. As traffic conditions can change rapidly due to accidents, road closures, or unexpected events, the system must recreate its prediction model each time a new prediction is required. By continuously updating the model with the latest traffic data, the system can offer accurate and up-to-date information to drivers, allowing them to choose the most efficient routes and alleviate congestion in real-time. 

By considering these restrictions, we come to the realization that the applicability of machine learning is greatly constrained. Many significant applications of learning methods in the past 50 years would have been impossible to solve without relaxing these restrictions. On the other hand, online learning assumes:~\cite{fontenla2013online}:
\begin{enumerate}
    \item only a portion of the data is available at any one time
    \item the response should be timely
    \item the data distribution can change over time
\end{enumerate}
\noindent

This study introduces a novel online linear regression model \textit{Online Regression with Weighted Average (OLR-WA)} that is based on the weighted average of a base model which represents the already seen data and an incremental model which represents the new data. OLR-WA eliminates the challenge of storage requirements for large amounts of data while providing an effective solution for large-scale problems. Additionally, it does not require any assumptions about the distribution of data and instead can work in an adversarial scenario, making it adaptable to a wide range of situations where the data may not be independently and identically distributed. 


This paper makes two significant contributions. First, the OLR-WA model performs comparably to a batch model over data consistent with the batch model expectations. Second, the OLR-WA model provides flexibility not found in the batch model and many other online models. 

\section{Related Work}

In this section, we review some of the work related to online learning and to linear regression. 

\subsection{Stochastic Gradient Descent (SGD)}
Gradient descent is a commonly used optimization algorithm for training machine learning models. It is based on the idea of iteratively adjusting the model's parameters in the direction of the negative gradient of the loss function, thereby minimizing the loss. Stochastic gradient descent (SGD) is a variation of gradient descent that uses a random sample of the data, rather than the full dataset, to compute the gradient at each iteration. This makes SGD more computationally efficient. SGD is one of the most widely used 
techniques for online optimization in machine learning \cite{bouchard2015online}. Incremental algorithms, like Stochastic Gradient Descent (SGD) have been found to be more effective on large data sets than batch algorithms, and are widely used ~\cite{jothimurugesan2018variance}. 
Mini-batch gradient descent combines the two approaches by performing an update for every
mini-batch of $n$ training examples.\cite{ruder2016overview}

SGD and mini-batch SGD are well suited to online learning and widely used in the industry \cite{bouchard2015online}. While SGD and mini-batch SGD can be used for online learning, it works under the assumption that all the observed data up to the present moment is consistently accessible. They process the data one sample or a small batch of samples at a time, updating the model parameters after each sample or batch. In addition to that, these methods suffer from different limitations like \cite{ruder2016overview} the sensitivity to the choice of the learning rate, which affects the convergence speed. A learning rate that is too small will result in slow convergence, while a learning rate that is too large may cause the algorithm to oscillate or not converge at all. Additionally, SGD might get stuck in a local minimum, which can result in poor predictions. 

\subsection{Linear Regression}
Linear Regression \cite{abu-mustafa2012} \cite{maulud2020review} is one of the most common and comprehensive statistical and machine learning algorithms. It is used to find the relationship between one or many independent variables and one dependent variable. It is a 
mathematical approach used to perform predictive analysis and can 
be used to determine causal relationships in some cases. 

Regression may either be simple or multiple regression.
Simple linear regression studies the relationship between two continuous (quantitative) variables. One variable, denoted `x', is regarded as the predictor, explanatory, or independent variable and the other variable, denoted `y', is regarded as the response, outcome, or dependent variable \cite{maulud2020review}. The model equation is represented by
$y= \beta_{0}  +  \beta_{1}x  + \epsilon$
Multivariate linear regression (MLR) is used to predict the result of an answer variable using a number of explanatory variables. The basic model for MLR is 
$y= \beta_0 + \beta_{1}x_{1}  + \cdots + \beta_{m}x_{m}  +   \epsilon$.
The formula to determine the formula matrix (usually called pseudo-inverse) is \cite{maulud2020review}
$$\hat{\beta} = (X^TX)^{-1}X^T\mathbf{y}$$ 
\noindent
where 
$\beta$= 
$\begin{bmatrix}
\beta_{0} \\
\beta_{1} \\
\vdots\\
\beta_{m}
\end{bmatrix}$, X= $\begin{bmatrix}
    1 & x_{11} & x_{12} & \ldots & x_{1m} \\
    1 & x_{21} & x_{22} & \ldots & x_{2m} \\
    \vdots & \vdots & \vdots & \vdots & \vdots\\
    1 & x_{n1} & x_{n2} & \ldots & x_{nm} 
\end{bmatrix}$, y = $\begin{bmatrix}
    y1\\
    y2\\
    \vdots\\
    y_n    
\end{bmatrix}$

\subsection{On-line Linear Regression Models}
In general, on-line learning models follow the framework in Algorithm~\ref{alg:generalonlineframework}, which aims to minimize the total loss incurred.

\begin{algorithm}
\caption{General Online Learning Framework\cite{Alex2008}}\label{alg:generalonlineframework}
\begin{algorithmic}[1]
\State Initialize $w_1$ = 0
\For{each round t=1,...,T:}
\State Get training instance $x_t$
\State Predict label $\hat{y_t}  \in \mathbb{R}$
\State Get true label $y_t \in \mathbb{R}$
\State Incur loss $l(\hat{y_t}, y_{t}, x_{t})$
\State Update $w_t$
\EndFor
\end{algorithmic}
\end{algorithm}

\subsubsection{Widrow-Hoff}
One of the on-line linear regression models is the Widrow-Hoff algorithm, also known as Least-Mean-Square (LMS) Algorithm. LMS combines stochastic gradient descent techniques with a linear regression objective function by considering data one point at a time. 
For each data point, the algorithm makes successive corrections to the weight vector in the direction of the negative of
the gradient vector. This eventually leads to the minimum mean square error. The LMS update rule is:
\begin{equation}
    w_{t+1} = w_{t} - \alpha (w_{t} . x_{t} - y_{t}) x_{t}
\end{equation}
where $\alpha$ is the learning rate, and $w_{t}$ is the weight vector of the current iteration.
This update rule has been derived so we can stay close to $w_t$, since $w_t$ embodies all of the training examples we have seen so far. In our work, we call this confidence bias, and can be modeled within OLR-WA \cite{fontenla2013online}\cite{mohri2018foundations}.

The Widrow-Hoff algorithm is commonly used adaptive algorithm due to its simplicity and good performance. However, the value of the learning rate parameter must be chosen carefully to ensure the algorithm converges. As an iterative algorithm, it can adapt to a rapidly changing data environments, but its convergence speed may be slower than other algorithms.
Additionally, the LMS algorithm has a fixed step size for each iteration and may not perform well in situations where the input signal's statistics are not well understood or bursty.\cite{fontenla2013online}

\subsubsection{Online Support Vector Regression}
Support Vector Regression (SVR) was introduced in the 1990s by Vadimir Vapnik and his colleagues (Drucker, Cortes, \& Vapnik, 1996) \cite{drucker1996support} while working at AT\&T Bell Labs. The detailed exploration of SVR can be found in Vapnik's book (Vapnik, 1999) \cite{vapnik1999nature}. Vapnik's SVR model distinguished itself from standard regression models by fitting a tube, commonly known as the $\epsilon$-Insensitive Tube, instead of a line. This tube, with a width denoted as $\epsilon > 0$, defines two sets of points: those falling inside the tube, which are considered $\epsilon$-close to the predicted function and are not penalized, and those falling outside the tube, which are penalized based on their distance from the predicted function. This penalty mechanism bears similarity to the penalization used by Support Vector Machines (SVMs) in classification tasks. In addition to the tube-based approach, SVR incorporates a kernel function. This kernel function allows SVR to capture nonlinear relationships between the input features and the target variable. It provides the flexibility to choose appropriate transformations for diverse types of data and problem domains.

The online variant of SVR employs stochastic gradient descent (SGD), which applies the concept of updating the dual variables incrementally based on the deviation between the predicted and target values. The dual variables, typically represented by $\alpha$, are optimization variables associated with each data point. These variables measure the importance or weights assigned to each data point in the training set. The values of the dual variables determine the influence of each data point on the final decision function. Data points with non-zero dual variables, referred to as support vectors, play a significant role in defining the decision boundary or regression surface.

Our approach fundamentally diverges from Online Support Vector Regression (Online SVR) in several key aspects. Firstly, Online SVR randomly selects one data point at a time from the entire pool of previously observed data using stochastic gradient descent. In contrast, our approach requires a minimum number of data points, forming a mini-batch, to construct a model. Moreover, our approach encompasses the ability to forget previously seen data points while preserving their associated metadata in the form of a weighted average generated model. This distinguishing feature of our approach proves advantageous, particularly in adversarial scenarios, as the weights can be tailored to favor user-specified criteria. By leveraging this weighted average model, our approach demonstrates flexibility and adaptability in such scenarios.

\subsubsection{Recursive Least-Squares
(RLS) algorithm}
The Recursive Least-Squares (RLS) algorithm is a type of adaptive filter algorithm that is used to estimate the parameters of an online linear regression model. It is a recursive algorithm that uses a least-squares criterion to minimize the error between the desired output and the estimated output of the system. 
In the context of online linear regression, the RLS algorithm is used to estimate the coefficients of the linear model. The algorithm starts with an initial estimate of the coefficients and updates them in real-time based on new data points. The algorithm uses a recursive update rule to adjust the coefficients based on the current data point and the previous estimates. The update rule is based on the gradient descent optimization method, which aims to minimize the mean square error between the desired output and the estimated output. 
The RLS algorithm uses a forgetting factor, which is a scalar value between 0 and 1, to balance the trade-off between the importance of the current data point and the importance of previous data points. A higher forgetting factor value gives more weight to the current data point, while a lower forgetting factor value gives more weight to previous data points. Similar weights are used in OLR-WA. In summary, the RLS algorithm is an efficient and robust algorithm that can estimate the parameters of online linear regression models in real-time. It has fast convergence and good performance in terms of stability and robustness. However, it can be computationally expensive when the number of input variables is large.\cite{fontenla2013online}

Our approach is fundamentally different from either LMS or RLS in that OLR-WA computes an incremental model based upon a collection of inputs and then integrates the incremental model into an existing base model representing all of the previous data. 
This allows model integration instead of data integration, which means larger bursts of data points can be handled and we can emphasize either the previous model, the incremental model or neither. 
While we intend to compare both performance and accuracy of OLR-WA with other techniques in the future, in this paper we validate OLR-WA by comparing it to linear regression with all data available.

\subsubsection{Online Ridge Regression}
Ridge regression is a regression technique that addresses the issue of overfitting in linear regression models by introducing a regularization term. Overfitting occurs when a model fits the training data too closely, resulting in poor performance on new, unseen data. Ridge regression adds a penalty term to the loss function during training to constrain the model's coefficients, thus reducing overfitting. \cite{van2015lecture} The cost function in ridge regression consists of two parts: the residual sum of squares (RSS) term $(y - Xw)^T(y - Xw)$ that measures the discrepancy between the predicted and actual values, and the regularization term $\lambda w^Tw$ that penalizes large coefficient values to prevent overfitting. The regularization term helps in controlling the complexity of the model and reducing the impact of irrelevant features.  \cite{mohri2018foundations} By applying stochastic gradient technique to ridge regression, we can derive a similar algorithm. In each round, the weight vector is updated with a quantity based on the prediction error $(y_t - X_tw_t)$. The main idea behind Online Ridge Regression is to update the model's parameters in an incremental fashion while incorporating regularization. This is achieved by adapting the standard ridge regression algorithm to handle streaming data. The difference between regular ridge regression and online ridge regression lies in the way the data is processed and updated. In online ridge regression, the data is processed sequentially, updating the coefficient vector $w$ after each observation. Here's the formula for online ridge regression:
\begin{equation}
\label{eqn:equationWW}
J(w)=(y - Xw)^T(y - Xw)+ \lambda w^Tw
\end{equation}
where $J(w)$ represents the cost function, $y$ represents the vector of observed or target values, $X$ represents the matrix of predictor variables or features, $w$ represents the coefficient vector or parameter vector, which contains the regression coefficients for each feature, $\lambda$ represents the regularization parameter, which controls the trade-off between fitting the training data and preventing overfitting, $(y - Xw)^T(y - Xw)$ represents the squared residual term, which measures the difference between the observed values $y$ and the predicted values obtained by multiplying the predictor variables $X$ with the coefficient vector $w$, and finally $\lambda w^Tw$ represents the regularization term, which penalizes the magnitude of the coefficient vector $w$ to prevent overfitting.

Our methodology takes a fundamentally different approach compared to Online Ridge Regression across several critical aspects. Firstly, whereas Online Ridge Regression utilizes stochastic gradient descent to randomly select individual data points from the entire pool of previously observed data, our approach requires a minimum number of data points to form a mini-batch for constructing the model. Furthermore, our method allows for the forgetting of previously encountered data points while retaining their associated metadata through the generation of a weighted average model. This distinctive feature of our approach offers clear advantages, particularly in adversarial scenarios, as the weights can be tailored to prioritize specific user-defined criteria. By leveraging this weighted average model, our approach demonstrates remarkable flexibility and adaptability in such challenging situations.

\subsubsection{The Online Passive-Aggressive (PA) algorithms}
The Passive-Aggressive algorithms is a family of algorithms for online learning. It is often used for classification tasks but can also be applied to regression. The algorithm updates the model's parameters in a way that minimizes the loss using the below update rule 
\begin{equation}
    w_{t+1} =  \arg \min_{w \in \mathbb{R}^n} \frac{1}{2}\lVert w - w_{t}\rVert ^2 \;\;\;\;\;\;\; s.t \;\;\;\;\; l(w;(x_{t},y_{t})) = 0
\end{equation}
while remaining \quotes{$\mathrm{passive}$} whenever the loss is zero, that is $w_{t+1} = w_{t}$ or \quotes{$\mathrm{aggressive}$} in which those rounds the loss is positive, then the algorithm aggressively forces $w_{t+1}$ to satisfy the constraint $l(w_{t+1};(x_{t},y_{t})) = 0$.
In addition to that, the algorithm has two other variations $PA- \uppercase\expandafter{\romannumeral1}$, and $PA- \uppercase\expandafter{\romannumeral2}$, which adds the terms $C\xi$, and $C\xi^2$ respectively. $C$ the \quotes{aggressiveness parameter} is a positive parameter which controls the influence of the slack term $\xi$ on the objective function. Larger values of $C$ imply a more aggressive update step. In other words, the parameter $C$ represents the regularization parameter, and denotes the penalization the model will make on an incorrect prediction.

By repeating the training process for each training example, the Online Passive-Aggressive algorithm adapts its parameters to minimize the loss while considering the aggressiveness of the updates. The aggressiveness of the updates allows the algorithm to quickly adapt to new patterns in the data. Overall, the Online Passive-Aggressive algorithm is a useful tool for online learning tasks, including online linear regression. It can adapt to changing data streams and make updates to the model's parameters based on the aggressiveness determined by the training examples encountered.\cite{crammer2006online}

The Passive Aggressive Online Algorithm is an efficient approach for learning on the fly in scenarios involving a continuous stream of data with labeled documents arriving sequentially. An illustrative use case is monitoring the entire Twitter feed 24/7, where each individual tweet holds valuable insights for prediction purposes. Due to the impossibility of storing or retaining all tweets in memory, this algorithm optimally processes each tweet by promptly learning from it and subsequently discarding it. The PA Online Algorithm, similar in some respects to the OLR-WA, possesses a mechanism for discarding data points. However, OLR-WA distinguishes itself from PA algorithm by incorporating weights that favor either the base or incremental model. However, PA algorithm exhibits a default behavior that leans towards favoring new incoming data points. 

\section{OLR-WA Methodology}

The OLR-WA algorithm creates a new linear regression model for each data sample. The sample model and the existing model are merged to form a new linear regression model which can be used to make predictions until the next sample arrives. We compare OLR-WA to a batch linear regression approach which waits until all of the data arrives to build a model. In addition to being able to make predicitions sooner, the final result of OLR-WA has similar results as to the batch model. 

\subsection{Data Sets}
In this study we use relatively small synthetic and real world data sets. Each synthetic data set is drawn from either a two dimensional data distribution or a three-dimensional data distribution. For our experiments, we consider three types of distributions. The first is where all of the data is from the same linear distribution with a low noise factor as variance. 
The second is also a linear distribution, but the variance is higher. The third data set is a combination of two data sets, each with different liner distributions. This is an adversarial situation representing a change in the data distribution. 
Each experiment is represented by 
the number of data points \quotes{$\mathrm{N}$}, the variance \quotes{$\mathrm{Var}$}, the correlation \quotes{$\mathrm{Cor}$} if it is positive, or negative, and the step size between data points \quotes{$\mathrm{Step}$}. the following figures show some distribution samples of the dataset with 2-D settings.

\begin{center}
\includegraphics[width=.3\textwidth]{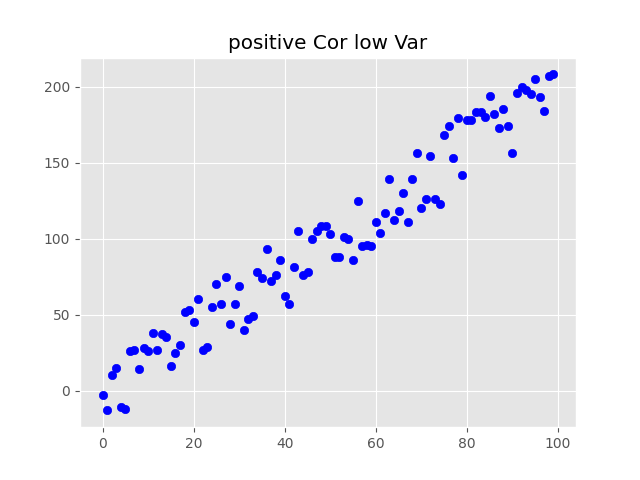}\hfill
\includegraphics[width=.3\textwidth]{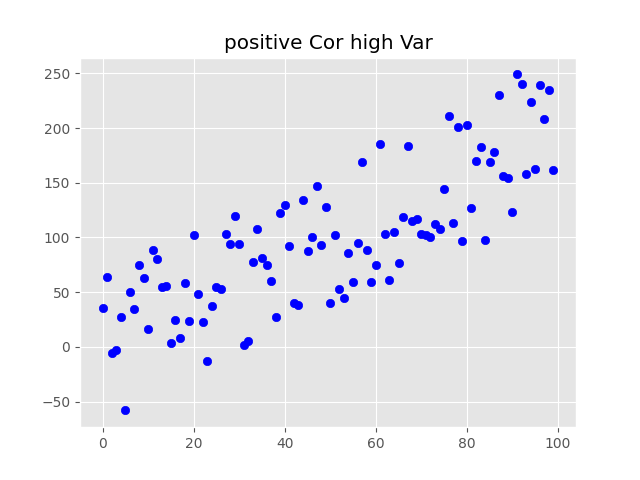}\hfill
\includegraphics[width=.3\textwidth]{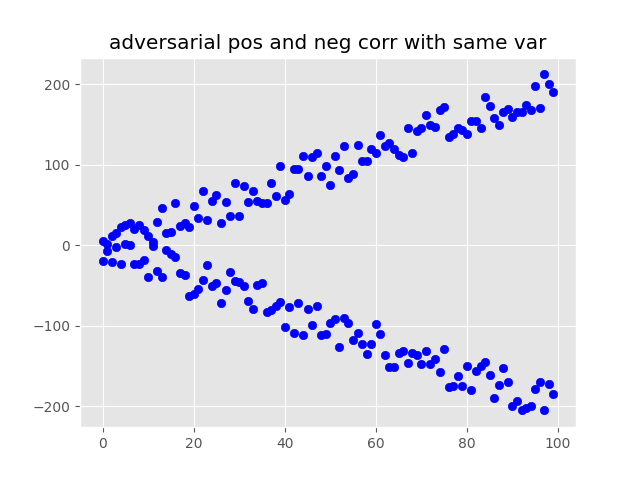}\hfill
\includegraphics[width=.3\textwidth]{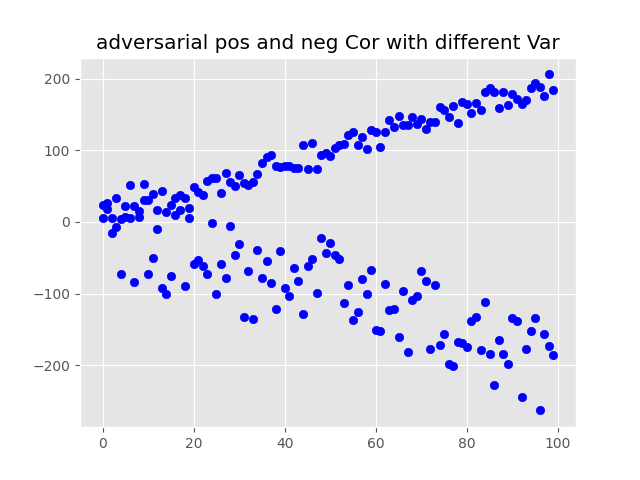}\hfill
\includegraphics[width=.3\textwidth]{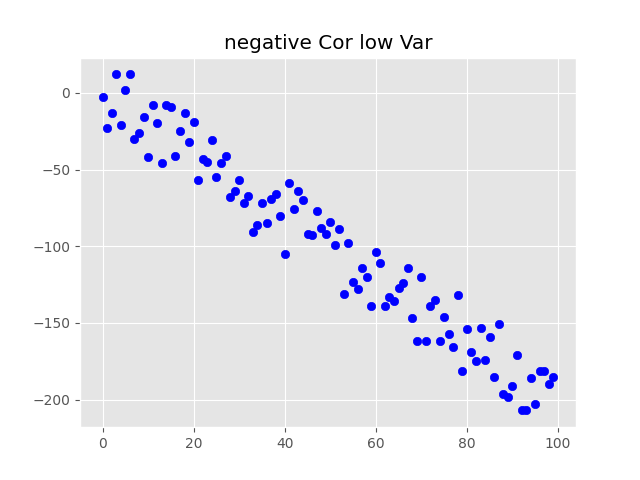}\hfill
\includegraphics[width=.3\textwidth]{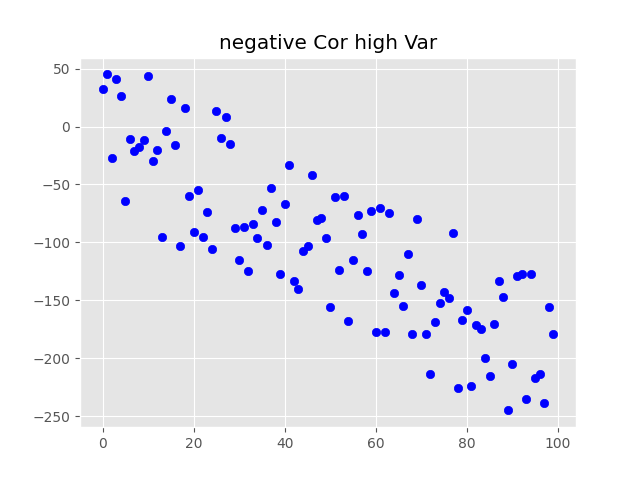}
\label{fig:datasetfigures}
\end{center}

In addition to that, for the purpose of validating our model with real world data sets, we used two public data sets explained in Section~\ref{Sec:Exp}. 

\subsection{Method}
Our approach includes two parts: a base model and an incremental model. The base model is the initial set of data points used as a starting point for linear regression. There is no specific limit to the number of data points, the minimum number necessary to create a model (e.g. 2 points for a 2-D model, 3 points for a 3-D model). The incremental model is a linear regression model that is created with each new set of data points. Again, there is no limit to the number of points to create a model, but we use the same guidelines as for the base model, (e.g. 2 points for a 2-D model, 3 points for a 3-D model). In our experiments, we allocate 10\% of the data points for the base model, while the remaining points are added incrementally at a rate of 10 points per increment.

Once both models are created, we calculate their weighted average. For 2D, this means averaging two lines, for 3D is averaging two planes and in higher dimensions, we must average hyperplanes.
The weighted average is computed by assigning user-defined weights to each model. This allows the user to modify the results based upon their knowledge of past and future data. If the incremental model is given higher weight, the model adapts to changing data more quickly. If the base model is given higher weight, the model is more resistant to transient changes. 
By default, the weights are equal, where both w-base and w-inc are assigned a fixed static number. Later in the paper, we will elaborate on various options for weight settings.

Although the techniques for computing the weighted average of lines, planes, or spaces may differ, the basic equation used remains consistent.

\begin{equation}
\label{eqn:equationX}
\textbf{V-Avg} =  (\textbf{w-base} \cdot  \textbf{v-base} + \textbf{w-inc} \cdot \textbf{v-inc})/(\textbf{w-base} + \textbf{w-inc})
\end{equation}

\noindent
where \textbf{w-base} represents the weight we assign to the base model, \textbf{w-inc} represents the weight we assign to the incremental model, 
\textbf{v-base} is the vector of the base model and \textbf{v-inc} is the vector of the incremental model. 
The weights here are scalar values, but the models will differ based on the number of dimensions. 
For example, in the 2D case, the base model is a line, so \textbf{v-base} will be computed using the equation
\begin{equation}
\label{eqn:equationXY}
   V = \langle x2-x1, y2-y1 \rangle  
\end{equation}
where (x1, y1) and (x2, y2) are the coordinates of the tail and head of the line respectively. 
Similarly, in 3D, if the normal vector of the plane equation $13x + 3y - 6z = 15$ then V = $\langle 13,3,-6 \rangle$ The same applies for \textbf{v-inc}.
\textbf{V-Avg} is the computed average vector, which we'll use later on with intersection point of the two models to construct the average line, plane, or space.

\begin{figure}[ht]
\centering
\includegraphics[width=0.6\textwidth,height=0.26\textheight]{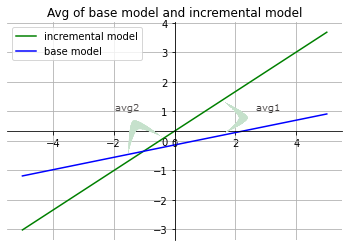}
\caption{Two computed averages}
\label{fig:fig4}.
\end{figure}

We can compute two weighted average vectors from the base model and the incremental model, Figure~\ref{fig:fig4} shows the idea in two-dimensional plane in which the linear regression models for the base and the increment are represented by lines. The formula generates two averages, labeled avg1 and avg2 in Figure~\ref{fig:fig4}. OLR-WA picks one of these two average vectors by detecting which one fits the data better. Since only the incremental data is available for evaluation, OLR-WA generates previous data from the base model. Using the incremental data and the generated data, the model selects the best fit. For example, in Figure\ref{fig:fig4} avg1 should be retained as the new base model, since it will have less mean square error. Algorithm~\ref{alg:cap} shows the steps of OLR-WA. 



\begin{algorithm}
\caption{Online Linear Regression with Weighted Average}\label{alg:cap}
\begin{algorithmic}[1]
\State base-regression = pseudo-inverse(base-X,base-y)
\For{$t$ $\gets$ 1 to $T$}
\State inc-regression = pseudo-inverse(inc-X,in-y)
\State \textbf{v-avg1} =  (\textbf{w-base} $\cdot$  \textbf{v-base} + \textbf{w-inc} $\cdot$ \textbf{v-inc})/(\textbf{w-base} + \textbf{w-inc})
\State \textbf{v-avg2} =  (\textbf{-1 $\cdot$ w-base} $\cdot$  \textbf{v-base} + \textbf{w-inc} $\cdot$ \textbf{v-inc})/(\textbf{w-base} + \textbf{w-inc})
\State intersection-point = get-intersection-point(base-regression, inc-regression)
\State space-coeff-1 = define-new-space(\textbf{v-avg1}, intersection-point)
\State space-coeff-2 = define-new-space(\textbf{v-avg2}, intersection-point)
\State err-v1= MSE(space-coeff-1)
\State err-v2= MSE(space-coeff-2)
\If{err-v1 $<$ err-v2}
    \State coefficients $\gets$ space-coeff-1
\Else
    \State coefficients $\gets$ space-coeff-2
    \EndIf
\EndFor
\State return coefficients
\end{algorithmic}
\end{algorithm}


We explore different techniques for assigning weights to data points, each with its own feasibility, correctness, and significance. Let's discuss each assignment technique along with short examples:

a) Time-based: \cite{cormode2009forward}, this technique assigns weights to data points based on their age or recency. The weight of a data point decreases as it becomes older. This method acknowledges that recent data points are more likely to be relevant to the current situation and gives them more importance. For instance, in a stock market prediction system, recent stock prices might carry more weight in determining the future trend compared to older prices.

b) Confidence-based: \cite{prasad2008decision}, this technique assigns weights based on the confidence or accuracy of the data point. Data points that are known to be more accurate or reliable are given higher weights. For example, in a sentiment analysis task, if certain labeled data points have been verified by experts or trusted sources, those points could be assigned higher weights due to their higher confidence level.

c) Fixed-based: \cite{younger1999fixed}, our model here offers the flexibility for two variations. The first variation assigns fixed equal weights to all data points, treating them equally in terms of importance. This implies that every data point contributes equally to the model. For instance, in a weather prediction model, each recorded temperature measurement might have the same weight. In this variation, the weight of the base model, w-base, is updated after each iteration by adding the weight of the incremental model, w-inc. This update is necessary because the current generated model incorporates both the base data and the incremental data, reflecting their combined influence. The second variation of the fixed-based technique assigns fixed equal weights to both the base model and the incremental model. This signifies that the existing state of the base model holds equal weight to the new model generated based on the incremental data. This ensures a balanced integration of the existing and new information. For example, in a machine translation system, both the previously trained model and the additional training data have equal importance in generating accurate translations.

One of the most common techniques is called the Decay Factor, which is a value used to reduce the weight of older data points in an online learning algorithm. The decay factor is multiplied by the weight of each data point at each iteration, so older data points have a lower weight than newer data points. This approach gives more importance to recent data points, as they are more likely to be relevant to the current situation.\cite{loshchilov2017decoupled} The decay factor is considered a time-based weighting scheme.
It is important to note that the choice of weighting scheme depends on the specific problem and dataset at hand, and it's important to experiment with different weighting schemes to find the one that works best for the specific use case. In the experiments section, we will demonstrate some techniques for adjusting weights to achieve the desired outcomes.

\subsection{Time Complexity}
The algorithm presented can be applied to any form of linear regression, but using the pseudo-inverse is a particularly common choice as it is polynomial time. We will be focusing on this approach in our analysis of the algorithm. The algorithm executes one pseudo inverse linear regression for the base model, and T pseudo inverse linear regression for incremental data, the pseudo inverse linear regression equation as we stated earlier is $\hat{\beta}$ = $(X^TX)^{-1}X^T\mathbf{y}$.

Given X to be a $M$ by $N$ matrix, where $M$ is the number of samples and $N$ the number of features. The matrix multiplications each require $O(N^2M)$,
while multiplying a matrix by a vector is $O(NM)$. Computing the inverse requires $O(N^3)$ in order to compute the LU or (Cholesky) factorization. 
Asymptotically, $O(N^2M)$ dominates $O(NM)$ so we can ignore that calculation. Since we're using the normal equation we will assume that M $>$ N,  otherwise the matrix $X^TX$ would be singular (and hence non-invertible), which means that $O(N^2M)$ asymptotically dominates $O(N^3)$. Therefore, the total complexity for the pseudo inverse is $O(N^2M)$. 

Our proposed algorithm uses a linear regression technique, specifically the pseudo-inverse linear regression, to process the data incrementally. This allows for the processing of smaller batches of data at a time, resulting in a lower overall data size. Compared to the traditional batch version of the pseudo-inverse linear regression, which has a time complexity of $O(N^2M)$, our online model's time complexity is likely to be $O(KN^2(M/K))$, where K is the number of iterations. Thus the total time complexity of OLR-WA is about the same as for the batch model running all at once.

\subsection{Evaluation Metric}
The coefficient  of  determination, \quotes{usually denoted by $R2$ or $r2$, is the proportion of variation of one 
variable (objective variable or response) explained by other variables (explanatory variables) in regression} \cite{kasuya2019use}. This is a widely used measure of the strength of the relationship in regression. It describes how well the model fits the data. An $r^2$ close to 1 implies an almost perfect relationship between the model and the data ~\cite{ozer1985correlation}. This coefficient is defined as~\cite{kasuya2019use}

\begin{equation}
\resizebox{0.5\linewidth}{!}{$r^2 = 1 - \dfrac{ \sum_{i=1}^{n}(y_{i} - \hat{y_{i}})^2}{ \sum_{i=1}^{n}(y_{i} - \overline{y}_{i})^2} = 1 - \dfrac{SE \; \hat{\mathbf{y}}}{SE \; \overline{ \mathbf{y}}}$}
\end{equation}

\noindent
where $\hat{\mathbf{y}}$ denotes the value of the objective variable (y) predicted by regression for the ith data point. The second term of this expression is the residual sum of squares divided by the sum of squares of y.

It is highly recommended to use the coefficient of determination as the standard metric for evaluating regression analyses in any scientific field because it is more informative and accurate than SMAPE, and does not have the interpretability limitations of other metrics such as MSE, RMSE, MAE, and MAPE \cite{chicco2021coefficient}. The coefficient of determination is used as the evaluation metric for our model.

\section{Discussion}
In this research, we carried out experiments utilizing both 2-dimensional and 3-dimensional datasets. We employed the versatility and management capabilities of our dataset generator to perform multiple experiments. We will now examine the model's behavior and implications.
\subsection{2-Dimensional}
\begin{figure}[ht]
\centering
\includegraphics[width=0.6\textwidth, height=0.3\textheight]{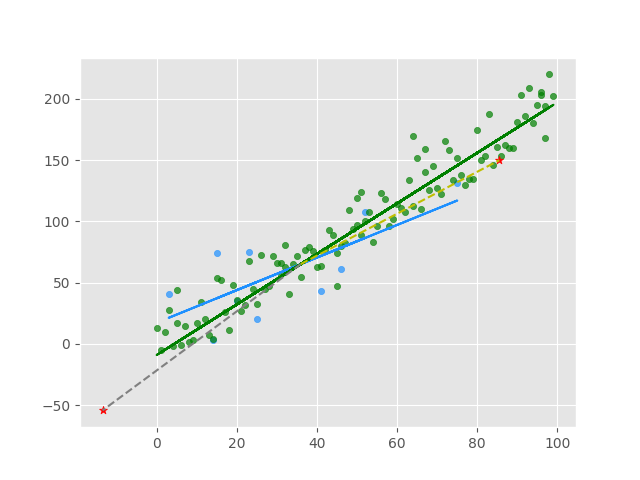}
\caption{Two average lines}
\label{fig:fig1}
\end{figure}

Figure \ref{fig:fig1}
provides a comprehensive illustration of a 2D scenario, the green points represent the base model points, which are used for visualization purposes only, but in reality, we don't retain them, the only thing we maintain about the those base points is the model itself which is represented here by a the the green line. The blue points are the new coming (incremental) points, the blue line represent the linear regression line for the incremental points. The yellow dashed line ends with a red star on its tail represents first computed average line, and the gray dashed line ends with a red star on its tail represents the second computed linear regression line.

In the 2D model, we define the average lines, the yellow and the gray in Figure \ref{fig:fig1} by the two norm vectors resulted from equation \ref{eqn:equationX} and a point of intersection between the base and the incremental lines. It is worth nothing that it is highly unlikely for the base and the incremental lines to have exactly the same slope, making them parallel, and thus having no intersection point. In this scenario, the current algorithm simply ignores the case and updates itself on the next iteration. Although this is a rare occurrence, for more accurate results, there are several solutions which we will consider in the future work.

As illustrated earlier, one of those two lines will be selected and will represent our current model, the other will be discarded. selection is based on the minimum Mean Square Error (MSE) of the new coming data and some sampled data from the base model.
\subsection{3 Dimensional}
\begin{figure}[ht]
\includegraphics[width=0.77\textwidth]{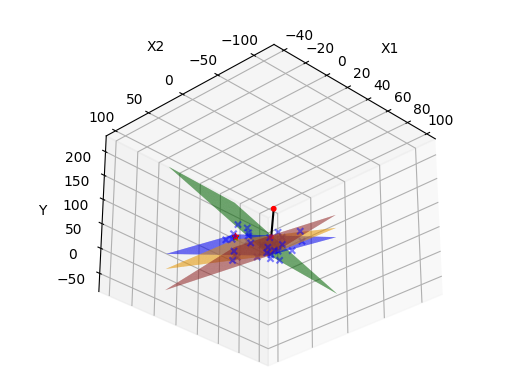}
\caption{Two average planes}
\label{fig:fig3}.
\end{figure}
Figure \ref{fig:fig3}
provides a comprehensive illustration of a 3D scenario, the blue plane represent the base model. The brown plane represents the incremental model. The orange plane represents first computed average plane, and the green plane represents the second computed plane. The 3D differs from the 2D with the coefficients. In 2D the coefficients are the slope or $m_{0}$, and the y\_intercept or (b), while in 3D the coefficients are $m_{0}$, $m_{1}$, $m_{2}$ or (b). In the 3D we define the average planes, the orange and the green in Figure \ref{fig:fig3} by the two norm vectors resulted from equation \ref{eqn:equationX} and a point of intersection between the base and the incremental plane. In this case, the intersection of two planes is a line, so we can use any point on the line. In general, the intersection point of two planes or spaces can be determined by solving the equation of both the base and incremental models, which will provide a point that is located within both planes or spaces.

\subsection{Generalization to Higher Dimensions (N-Dimensional)}
While OLR-WA is straightforward in 2-D and 3-D, the same procedure can be applied to higher dimensions. we only need to find a point in the intersection and the direction vector, and using those two elements, we can compute the new hyperplane. Obtaining the point of intersection differs from one dimensional space to another. For example, in 2-D, finding the intersection of two lines requires solving two equations with two variables, which results in an exact one point. However, in 3-D, finding the intersecting line required solving two equations with three variables, which produces an infinite number of solutions which are points representing that line. In higher dimensions, the intersection of N-Dimensional hyperplanes, is a (N-1)-Dimensional hyperplane which will be generated by solving two equations of N variables. This can be done using methods like Gaussian elimination, matrix inversion, or using software packages. The solution will be a point $(x_1, x_2, ..., x_n)$ that satisfies both equations, representing the intersection of the two hyperplanes. Note that there can be multiple solutions, depending on the hyperplanes' orientation and position in the N-dimensional space.

\subsection{Experiments}
\label{Sec:Exp}

We have conducted several experiments using 2-D and 3-D, we used 10\% of the total points as the base model, and a total of 10 points on each iteration for the incremental model. In the following we will provide a summary of these experiments and the outcomes that were obtained.

\subsubsection{Experiment 1}
In this experiment we generated 5 different datasets of 200 random, positively correlated points.  The $r^2$ for the batch model ranged from 0.9152 (in 3D) to 0.9558 (also in 3D), while the $r^2$ for OLR-WA ranged from 0.8991 (in 2D) to 0.9517 (in 3D). Tables 1a and 1b show the sample runs of the experiment.

\subsubsection{Experiment 2}
In this experiment, we generated 5 different datasets of 200 random, positively correlated points. However, this time 100 data points are randomly generated with one variance and the other 100 data points are generated with a different variance. Not surprisingly, neither model was able to model the data as accurately as Experiment 1, with the $r^2$ for the batch model ranging from 0.8222 (in 3d) to 0.9190 (in 2D). while OLR-WA $r^2$ ranges from 0.8107 (in 3D) to 0.8904 (in 2D). Tables 2a and 2b shows the sample runs of the experiment.

\noindent

\begin{table}[ht]
    \begin{subtable}[ht]{0.45\textwidth}
        \centering
        \begin{tabular}{||c c c c||} 
        \hline
        Experiment & Batch & Online & \\ 
        \hline\hline
        1 & 0.9348	& 0.9331 & \\
        2 & 0.9492	& 0.9349 & \\
        3 & 0.9379	& 0.8991 & \\
        4 & 0.9388	& 0.9291 & \\
        5 & 0.9250	& 0.9134 & \\ 
        \hline
        \end{tabular}
    \caption{Batch vs. Online for Positively Correlated Data}
    \label{Table2DCons}
\end{subtable}
\hfill
    \begin{subtable}[ht]{0.45\textwidth}
        \centering
        \begin{tabular}{||c c c c||} 
        \hline
        Experiment & Batch & Online & \\ 
        \hline\hline
        1 & 0.8444 & 0.8340 & \\
        2 & 0.8710	& 0.8323 & \\
        3 & 0.8982	& 0.8869 & \\
        4 & 0.9190	& 0.8904 & \\
        5 & 0.8795	& 0.8695 & \\ 
        \hline
        \end{tabular}
    \caption{Batch Vs. Online for Dataset with Shifting Variance}
    \label{Table2DShift}
\end{subtable}
\caption{Results of 2D Experiemnts}
\end{table}

\begin{table}[ht]
    \begin{subtable}[ht]{0.45\textwidth}
        \centering
        \begin{tabular}{||c c c c||} 
        \hline
        Experiment & Batch & Online & \\ 
        \hline\hline
1 & 0.9348 & 0.9326 & \\
2 & 0.9267 & 0.9080 & \\
3 & 0.9558 & 0.9517 & \\
4 & 0.9152 & 0.9156 & \\
5 & 0.9422 & 0.9449 & \\
        \hline
        \end{tabular}
    \caption{Batch vs. Online for Positively Correlated Data}
    \label{Table3DCons}
\end{subtable}
\hfill
    \begin{subtable}[ht]{0.45\textwidth}
        \centering
        \begin{tabular}{||c c c c||} 
        \hline
        Experiment & Batch & Online & \\ 
        \hline\hline
1 & 0.8400 & 0.8280 & \\
2 & 0.8738 & 0.8591 & \\
3 & 0.8222 & 0.8209 &  \\
4 & 0.8483 & 0.8470 & \\
5 & 0.8259 & 0.8107 & \\
        \hline
        \end{tabular}
    \caption{Batch vs. Online for Data with Shifting Variance}
    \label{Table3DShift}
\end{subtable}
\caption{Results of 3D Experiments}
\end{table}

\noindent
\subsubsection{Experiment 3}
The aim of the this experiment is to expose OLR-WA represented by algorithm~\ref{alg:cap} into real world datasets and validate its performance, we used two public datasets:

\begin{enumerate}
    \item \cite{dataset_ref1} 1000 companies data set. The dataset includes sample data of 1000 startup companies operating cost and their profit. Well-formatted dataset for building ML regression pipelines. This  data set is used to predict profit from R\&D spend and Marketing spend
    \item \cite{Cortez:2008:DMPS} Math Student data set. This is a dataset from the UCI datasets repository. This dataset contains the final scores of students at the end of a math programs with several features that might or might not impact the future outcome of these students. Math Student dataset is used to predict secondary school student's performance (final grade) using first period grade, and second period grade.    
\end{enumerate}

Tables~\ref{TableRealdataSet1} and~\ref{TableRealDataSet2} show the performance of OLR-WA versus the standard batch model using the 2 aforementioned data sets.
\begin{table}[ht]
    \begin{subtable}[ht]{0.42\textwidth}
        \centering
        \begin{tabular}{||c c c c||} 
        \hline
        Experiment & Batch & Online & \\ 
        \hline\hline
1 & 0.8078 & 0.7420 & \\
2 & 0.8151 & 0.8001 & \\
3 & 0.9979 & 0.9968 & \\
4 & 0.6972 & 0.5949 & \\
5 & 0.9976 & 0.9417 & \\
        \hline
        \end{tabular}
    \caption{Batch vs. Online for \cite{dataset_ref1} 1000 Companies Data Det.}
    \label{TableRealdataSet1}
\end{subtable}
\hfill
    \begin{subtable}[ht]{0.42\textwidth}
        \centering
        \begin{tabular}{||c c c c||} 
        \hline
        Experiment & Batch & Online & \\ 
        \hline\hline
1 & 0.8027 & 0.7913 & \\
2 & 0.7829 & 0.7818 & \\
3 & 0.8453 & 0.8448 &  \\
4 & 0.8302 & 0.8118 & \\
5 & 0.8802 & 0.8759 & \\
        \hline
        \end{tabular}
    \caption{Batch vs. Online for \cite{Cortez:2008:DMPS} Math Student Data Set }
    \label{TableRealDataSet2}
\end{subtable}
\caption{Results of Experiments on Real Data Sets.}
\end{table}
\noindent
\subsubsection{Experiment 4}
In this experiment we show an adversarial scenario. As in Experiment 2, we generate 200 random data points. However, in this case, the first 100 points are positively correlated while the next 100 points are negatively correlated. As expected, the batch model does not generate strong $r^2$ results (and it can be argued that is correct), see Figure \ref{fig:fig8}. However, in the online approach, the user can supply weights to indicate their preference for the data. For example, older points can be given higher, lower or the same weight as newer points. We classify the weights as \textbf{time-based} in which older points have lower weights, \textbf{confidence-based} in which older points have higher weights and \textbf{fixed-based} in which the relative weights are established a priori.  

\begin{figure}[!htb]
\minipage{0.45\textwidth}
\includegraphics[width=\linewidth]{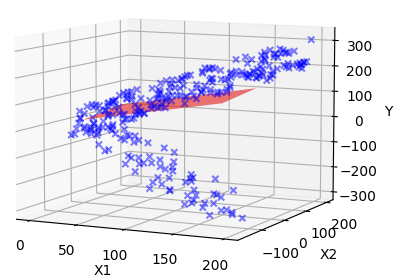}
  \caption{Fixed-Based Weights - equal points weights}\label{fig:fig20}
\endminipage\hfill
\minipage{0.45\textwidth}
\includegraphics[width=\linewidth]{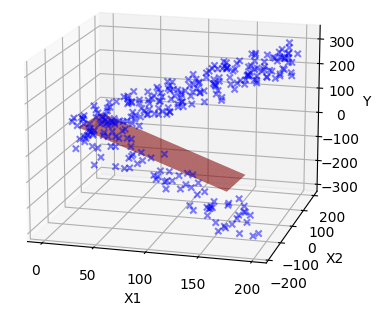}
  \caption{Fixed-Based Weights - equal models weights}\label{fig:fig21}
\endminipage\hfill
\minipage{0.45\textwidth}
\includegraphics[width=\linewidth]{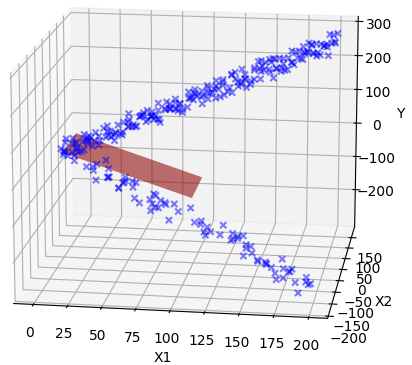}
  \caption{Time-Based Weights}\label{fig:fig5}
\endminipage\hfill
\minipage{0.45\textwidth}%
  \includegraphics[width=\linewidth]{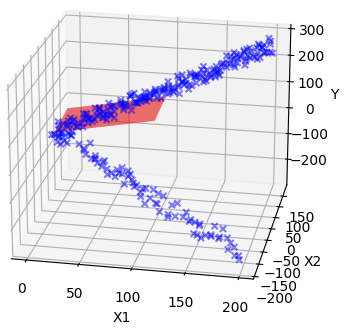}
  \caption{Confidence-Based Weights}\label{fig:fig7}
\endminipage
\end{figure}

\textbf{Fixed-based weights}
The default case, known as the fixed-based weights case, presents two options for the user's consideration. Firstly, all points are assumed to be treated equally, resulting in equal weights for each point. Secondly, both models are assigned fixed equal weights throughout the process. We will delve into these two cases in the following sections:

In the first case, assuming equal weights for all points, the base model begins with 40 points, denoted as w-base=40. Meanwhile, the incremental model processes 10 points during each iteration, denoted as w-inc=10. After each iteration, the weight of the base model increases as it accumulates the incremental points, given by the equation w-base += w-inc. Consequently, the base model progressively carries a higher weight than the incremental model. As a result, the model's regression plane aligns with the base model, as illustrated in Figure~\ref{fig:fig20}.

In the second case, the user assigns fixed equal weights to both models from the outset. Specifically, the user designates w-base = 1 and w-inc = 2. In this scenario, the regression plane gradually shifts towards the incremental model and aligns with it, as depicted in Figure~\ref{fig:fig21}.

\textbf{Time-Based Weights}
Figure~\ref{fig:fig5} shows the same distribution (but different data points), but this time the incremental model has a weight 20 times greater than the existing model. This represents a scenario in which the model trusts the data will continue with the new distribution. Note that the plane is a good fit for the incremental model. 

\textbf{Confidence-base Weights}
Figure~\ref{fig:fig7} again has the same data distribution, but different data points. In this case, the weight of the existing model is 20 times higher than the weight of the incremental model. This scenario entails placing greater trust in the initial data while assigning minimal weight to the incremental data in order to diminish its impact. Note that the plane is a good fit for the existing model. 

Between the three models, the time-base approach is the most dynamic. For example, a 3rd increment with yet another distribution would cause the model to change to match the new data, while the fixed weights would change less and the confidence weights the least.

\section{Conclusion and Future Work}
\begin{figure}[ht]
\centering
\includegraphics[width=0.60\textwidth]{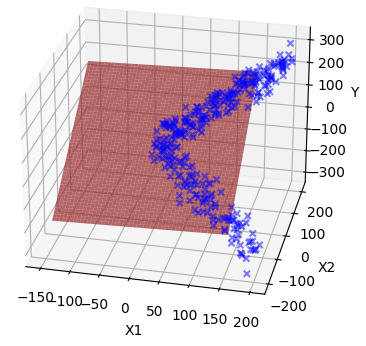}
\caption{Standard Batch Model with Adversarial Scenario}
\label{fig:fig8}.
\end{figure}

We have conducted 2D, and 3D experiments using OLR-WA and a batch model. In general, OLR-WA preformed comparable to the batch model. For example, in the 2D case with constant variance, the $r^2$ values for the batch model varied from 0.9251 to 0.9492 while the $r^2$ for OLR-WA varied from 0.8991 to 0.9349. Similar results hold for the experiments with shifting variance and the 3D experiments. Furthermore, the time complexity of OLR-WA is on par with the batch version. Additionally, it provides flexibility that is not a standard part of the batch model by allowing dynamic adjustment of weights. In other words, the standard batch model lacks the flexibility required to handle adversarial scenarios like OLR-WA. For instance, as illustrated in Figure \ref{fig:fig8}, when incremental data has a significant shift that calls for a new model, the $r^2$ of the standard batch model will be considerably low.



The work can be extended in many ways. First, while the model should adapt to any dimensionality, we have not tested extending the implementation beyond 3D. Second, while the time complexity of the model is on par with the batch model, we have not run performance tests to determine the impact of incremental evaluation, especially with respect to very large data sets. Third, the data sets were designed to find good fits. We need to compare the batch method with OLR-WA on datasets with significantly greater noise (much lower $r^2$ values).  One issue specific to OLR-WA that can be addressed is the case where the two models do not intersect. One option is finding the closest distance between the two lines that can be used instead or possibly increasing the incremental model size to resolve the problem of parallelism.

While OLR-WA shows promise when compared to the batch model, we also need to compare it to other incremental learning techniques, such as LMS and RLS. The comparisons need to consider data distributions, adversarial situations and large data sets (both in terms of the number of features and the number of points).

Finally, OLR-WA has potential for interesting extensions, specifically in the area of weight selection. Currently, weights (w-base, and w-inc) are chosen by the user, but it would be fascinating to give OLR-WA the ability to select weights based on both user preferences and/or observed data. By recording data such as the number of points, variance, and correlation of the base model and each incremental model, as well as the user's preferred weight classes (time, confidence, or fixed), OLR-WA can automatically select weights. 
For instance, if the user favors time-based weights and the incremental model significantly deviates from the existing model, OLR-WA could automatically increase the weight of w-inc which represents the incremental data weight. Likewise, when presented with adversarial data, OLR-WA could automatically increase the size of the incremental batch or consider many mini-batches to allow more points to accumulate and determine whether the new data fits a different model or represents an outlier, if it is highly likely that the adversarial data is constituting a new different model, and the user favors the weight-time weights, then, OLR-WA automatically will increase w-inc to the favor on the new model, and similarly if the user favors the base model to a certain extent, OLR-WA will automatically increase w-base which represents the base model weight till that point in time. 
Another example will be computing the incoming tweet or post weight based on the likes or positive/negative comments it receives. Furthermore, introducing a forgetting factor can enhance this capability. When the model detects the introduction of a new model through a series of incremental mini-batches, it can be programmed to forget the old model and begin considering the new model. This approach increases the flexibility of the model and enables it to adapt to changing data distributions more effectively.

%
%
%
\bibliographystyle{splncs04}
\bibliography{bib}
\end{document}